\documentclass{article}

\usepackage{times}
\usepackage{graphicx} 
\usepackage{subfigure} 

\usepackage{amsmath}
\usepackage{amssymb}
\usepackage{url}

\usepackage{etoolbox}
\usepackage{natbib}

\usepackage{algorithm}
\usepackage{algorithmic}

\usepackage{hyperref}

\usepackage[accepted]{icml2017} 

\icmltitlerunning{Automatic Representation for Lifetime Value Recommender Systems}

\begin{document} 
	
	\twocolumn[
	\icmltitle{Automatic Representation for Lifetime Value Recommender Systems}
	
	
	
	
	\begin{icmlauthorlist}
		\icmlauthor{Assaf Hallak}{MSR}
		\icmlauthor{Yishay Mansour}{MSR}
		\icmlauthor{Elad Yom-Tov}{MSR}
	\end{icmlauthorlist}
	
	\icmlaffiliation{MSR}{Microsoft ILDC, Herzeliya, Israel}
	
	\icmlcorrespondingauthor{Assaf Hallak}{ifogph@gmail.com}
	
	
	\vskip 0.3in
	]
	
	
	
	\printAffiliationsAndNotice{} 

\newcommand\given[1][]{\:#1\vert\:}


\begin{abstract}
Many modern commercial sites employ recommender systems to propose relevant content to users. While most systems are focused on maximizing the immediate gain (clicks, purchases or ratings), a better notion of success would be the lifetime value (LTV) of the user-system interaction. The LTV approach considers the future implications of the item recommendation, and seeks to maximize the cumulative gain over time. The Reinforcement Learning (RL) framework is the standard formulation for optimizing cumulative successes over time. However, RL is rarely used in practice due to its associated representation, optimization and validation techniques which can be complex. In this paper we propose a new architecture for combining RL with recommendation systems which obviates the need for hand-tuned features, thus automating the state-space representation construction process. We analyze the practical difficulties in this formulation and test our solutions on batch off-line real-world recommendation data. 
\end{abstract}

%
%

\section{Introduction}
Recommender systems (RS) study the problem of optimizing the user interaction with the item (product) catalog. There are two possible setups: in the explicit setup, user feedback is given in rating for a subset of items from the catalog. In the implicit setup the user's preferences can be inferred from his history of acceptance or rejection of items. For both settings, the recommender's goal in each interaction is to recommend new items that will have high rating (explicit setup) or will be accepted with high probability (implicit setup). Recommender systems often interact with the same user repeatedly, and seek to improve the recommendations through personalization and cross-user inference. RS formulation have been applied to several commercial domains including movies, music and games recommendations, as well as some marketing schemes \cite{resnick1997recommender,adomavicius2005toward,ricci2011introduction}.

Traditionally recommender systems learn the preference of each user from offline batch data. The learned model is then used to recommend additional items when the user queries the system. As more data is collected, the model is refined according to the new samples. One specific approach related to our work is Matrix Factorization (MF; \cite{koren2009matrix}) which learns the user-item preference matrix through partial or noisy observations. Another related formulation, more suitable for rapidly changing environment and users with few interactions is contextual multi-armed bandits \cite{li2010contextual} in which recommendations are given on-line based on the user's context (comprised of aggregated user-information). These suggested solutions' aim is to optimize the success probability of the current interaction with the user. However, for systems that interact with the same users repeatedly over time, a more suitable metric would be the cumulative successes over time, also known as lifetime value (LTV) \cite{theocharous2013lifetime, pfeifer2000modeling}. The LTV perspective can be beneficial in many scenarios that myopic policies (optimizing the current interaction result) might find challenging:

\begin{itemize}
	\item \textbf{Expanding the user's taste} -- For example, recommending a song of a non-mainstream singer, if accepting the recommendation will likely pique the user's interest in that singer's other songs.
	\item \textbf{User sensitivity} -- Some recommendation are prone not only to fail, but also to offend the user due to their personalized content. For example recommending a weight loss, elderly reading aid, or a dating app. While these apps might have high acceptance probability, an insulted user might avoid further interaction with the recommendation service.
	\item \textbf{Sequels and references} -- Users are more likely to accept items in their natural order. For example, the third "Lord of the Rings" might be the best of its series, but a good recommender presents the episodes in sequential order. A less obvious example can be seen in movies that reference older movies such as parodies and remakes.
	\item \textbf{Self preservation} -- Consider a recommender system that suggests a competitor to users. If the recommendation is accepted, it can lead to losing the customer. Similar instances can manifest quite commonly: ads that forward the user to a different domain, games that engage the user for a long time and produce no revenue, or movies that motivate the user to do sports. 
\end{itemize}

The common principal in these examples is that optimizing over the recommendations in the current interaction without considering the future impact is sub-optimal and can even result in a substantial future loss. The alternative view which considers the dynamics of the interaction and optimizes over the cumulative successes (or LTV) is usually formalized through the Reinforcement Learning (RL) framework using Markov Decision Processes (MDPs) \cite{sutton1998reinforcement}. 

While the idea of optimizing interaction with the user over time has been around for many years \cite{jonker2004joint, pednault2002sequential, pfeifer2000modeling}, the formulation is usually content based which leads to hand-tuned features and dynamics; These rarely provide an adequate representation of the user at a given time. One different solution was proposed in \cite{shani2005mdp}, where the aggregated past $k$ recommendations are used to define the current state. This approach has scalability issues and cannot cope with high $k$ values or with a large recommendations set.

In this paper we propose the automatic construction of features for RL which aggregates all past recommendations to one feature vector using the MF framework. We follow a common scenario in which maximizing the LTV is considered as an improvement for a given recommender, and lay down the workflow for answering whether it should be incorporated or not.


\section{Matrix Factorization} \label{Sec:MF}
Our automatic feature construction is based on the MF technique commonly used for RS. Assume there are $m$ users and $n$ items, MF-algorithms receive as input the sampled rating matrix $\mathcal{R} \in \mathbb{R}^{m,n}$ which contains in the $(i,j)$ coordinate the rating user $i$ gave item $j$ (or $0$/$1$ if the item was rejected/accepted). If the item was not rated by the user a default value is used (commonly $0$, but can also be the average rating in the set). The algorithm outputs $k$-dimensional latent vectors for each user and item written in matrix form by $U \in \mathbb{R}^{m,k}$ and $V \in \mathbb{R}^{k,n}$ correspondingly. These vectors are then used to reconstruct the rating matrix $F$ through various models. For example, the simplest method solves: 
\begin{equation} \label{Eq:Als}
U V=\mathcal{R}, \quad \quad U \in \mathbb{R}^{m,k}, V \in \mathbb{R}^{k,n}.
\end{equation}
Minimizing the squared difference between the observed samples and the current estimates of $U, V$ can be done using stochastic gradient descent or alternating least squares, where regularization and bias terms can be added to improve the model or reduce over-fitting \cite{koren2009matrix}.

More complex approaches include Bayesian Personalized Recommendation (BPR; \cite{rendle2009bpr}), Probabilistic Matrix Factorization (PMF; \cite{salakhutdinov2008bayesian}) and Poisson Factorization (PF; \cite{gopalan2014bayesian}) that motivate the model through a Bayesian perspective,  CLiMF \cite{shi2012climf} that maximizes a ranking based score, and matrix completion techniques that find a low rank representation using nuclear-norm based optimization and singular value decomposition (SVD) \cite{gogna2015svd}. 

We propose an architecture in which any MF algorithm can be plugged in as a black-box. However, our architecture requires multiple user queries after the construction of the model, so in terms of complexity, queries should be inexpensive. This does not mean a better MF algorithm should not be preferred - on the contrary, improved success rate of the myopic recommendation strategy is likely to improve the LTV cumulative success as well. 

\section{Reinforcement Learning} \label{Sec:MDP}
Reinforcement Learning (RL) is an umbrella term for a specific set of problems in the machine learning community dealing with sequential interaction of an agent with a stochastic environment. In this section we provide the key notations and definitions required for the paper; For more details on RL we refer the reader to \cite{sutton1998reinforcement, bertsekas1995dynamic}.

An MDP is a tuple $M=(\mathcal{S}, \mathcal{A}, P, R, \nu)$ where $\mathcal{S}$ is the state space, $\mathcal{A}$ is the action space, $P$ is the transition probability kernel, $R$ is the reward function and $\nu$ is the initial state distribution. The process iterates as follows: first an initial state is sampled from $\mathcal{S}$: $s_0 \sim \nu$. Afterwards, each time step $t=0,1,...$ the agent chooses an action $a_t \in \mathcal{A}$, receives a stochastic reward $r_t \sim R(s_t, a_t)$ and the state transitions to $s_{t+1} \sim P(s_t, a_t)$. A strategy for choosing actions given the current state $s$ is called a policy, we denote by $\pi(a|s)$ the probability to choose the action $a$ in state $s$ according to policy $\pi$. 

Due to the everlasting nature of recommender systems, we consider the $\gamma$-discounted infinite horizon setup. The LTV of a specific state $s$ (also called the value function) when following policy $\pi$ is the expected cumulative discounted reward starting from state $s$: $V^\pi(s) = \mathbb{E}_\pi \left[ \sum_{t=0}^\infty \gamma^t r_t \given[\Big]  s_0 = s \right]$, with mean $J^\pi = \sum_{s\in \mathcal{S}} \nu(s) V^\pi(s)$. The problem of finding the policy maximizing $J^\pi$ is called planning. We also denote the $Q$-function: $Q^\pi(s,a) =  \mathbb{E}\left[ R(s,a) \right] + \gamma \sum_{s' \in \mathcal{S}} P(s'|s, a) V^\pi (s')$, which is the cumulative expected value of first taking action $a$ in state $s$, and then following policy $\pi$ for the rest of the actions. The Q-function is often used to make a one-step improvement of the current policy by maximizing its value over the actions.

When dealing with problem that has an excessively large state space, the common practice is to find features for every state $\phi(s) \in \mathbb{R}^d$ where $d$ is the number of features, and then use a linear function approximation for the value function: $V(s) \approx \theta^\top \phi(s)$ for some weight vector $\theta \in \mathbb{R}^d$. Similarly, if the action space is large the $Q$-function can be approximated through $Q(s,a) \approx \theta^\top \phi(s,a)$ for some other feature vector $\phi(s,a) \in \mathbb{R}^d$.

A key problem in the RL field is called policy evaluation: Given batch data, find the value $V^\pi(s)$ for a specific policy $\pi$ which may have generated the data (on-line policy evaluation), or not (off-line policy evaluation). With linear function approximation, this translates to finding an appropriate weight vector $\theta$. Fortunately, there are already several solutions in the literature for solving the policy evaluation problem which will be discussed in Section \ref{Sec:Off-line}.

\section{Model Construction} \label{Sec:Model}
To apply techniques from the RL literature, we require a formulation of a dynamic problem consisting of (1) sequential states $s_t \in \mathcal{S}$ where $\mathcal{S}$ is the state space, (2) sequential actions $a_t \in \mathcal{A}$ where $\mathcal{A}$ is the action space and (3) a reward signal $r_t$.

In sequential recommender systems, the state $s_t$ relates to some inner unknown state of the user at time $t$. Hence, if there are many users, each user interaction represents a different state trajectory. The action $a_t$ is naturally chosen to be the recommended item at time $t$, so the action space $\mathcal{A}$ is the item catalog. The reward signal $r_t$ is the rating the user gave to item $a_t$ at time $t$, or $0$/$1$ for implicit ratings.

Since the actual state of the user at each time step $s_t$ is not observed directly, its feature representation $\phi(s_t)$ is used instead. Features are often composed of engineered parameters such as the time steps since the last click, the cumulative reward or the user's demographics \cite{pednault2002sequential,theocharous2015ad}. However, these suffer from several disadvantages:
\begin{itemize}
	\item Scaling and tuning -- A number of feature engineering techniques have been applied to improve recommendation quality (e.g. log transformation, clipping, normalization). It is non-obvious if and when these techniques should be applied, as mishandling can cause good algorithms to perform poorly. 
	\item Hybrid features -- Binary (e.g. gender), categorical (e.g. country), discrete (e.g. number of interactions) and continuous (e.g. rate of success). Different types of features composed together in a single vector require more human intervention and adaptation. 
	\item Redundancy - Many system-generated features provide a poor representation of the user (e.g. operating system of the user), other features subsets give similar information on the user (e.g. timezone and location). These features unnecessarily increase the complexity of the problem.
	\item Trivial dynamics -- most engineered features advance trivially with time, for example, ``user gender'' is a static feature, and ``cumulative rewards'' is a simple counter. This suggests the essence of the process is not captured well in the dynamic.
\end{itemize}
While the points above are concerned with the state space representation, it is not uncommon for an RS to provide recommendations from a large item catalog corresponding to a large action space\footnote{The action space is usually more sensitive to size - most RL benchmarks consider a few actions (Mountain Car, Acrobot, Atari games, etc. \cite{duan2016benchmarking}), while in RS there are usually thousands of possible recommendations.}. Hence a feature vector for a state-action pair is needed, and not just a feature vector for every state. Our main goal in this paper is to automate the feature generating process. 

The proposed solution relies on the Matrix Factorization (MF) technique for RS which encodes each user and each item using a latent vector. These vectors can be considered to contain the relevant information for assessing the success probability of each user-recommendation interaction, and so we propose using them as the basis for state space and action space feature construction. 

Specifically, assume we are given such an MF black-box that given batch data of the form (User ID, Item ID, Reward signal, Timestamp), learns a model fitting every item and every user with their latent vectors of dimension $k$. In addition, we expect this model to support queries composed from a user's $i$ history of system interactions $h_i$: $u_i=\text{MF} \left(h_i \right)$, which returns the corresponding latent vector for this user. We propose the following architecture:

\textbf{Input}: Off-line batch data in the form of (User ID, Item ID, Reward, Timestamp), dimension of latent space $k$.
\begin{enumerate}
	\item Apply an MF algorithm on the data such that each user $i$ and item recommendation $j$ are coupled with the corresponding vectors $u_i, v_j \in \mathbb{R}^k$. From here on items are referred to by their vector representation $\{ v_j \}_{j=1}^n$.
	\item For each user $i$, aggregate the data to trajectories: $\left(v_{i,t}, r_{i,t} \right)_{t=0}^{T_i}$ using the resulting representation and the timestamp, where $T_i$ is the number of interactions with the user. Denote the $i$'th user's history at each time step $t=0..T_i$: $h_{i,t} =  \left(v_{i,l}, r_{i,l} \right)_{l=0}^{t} $.
	\item For each user trajectory, generate the state feature trajectory as follows: $\phi(s_{i,t}) = \text{MF} (h_{i,t})$ for every $t=0..T_i$.
	\item Apply any RL technique on the resulting $m$ trajectories $\{ \left( \phi(s_{i,t}), v_{i,t}, r_{i,t} \right)_{t=0}^{T_i} \}_{i=1}^m$. 	
\end{enumerate}
This way, by using the latent representation of the recommendation space, we relieve the practitioner from engineering hand-picked features. 

\begin{figure}  
  \centering
    \includegraphics[width=0.48\textwidth]{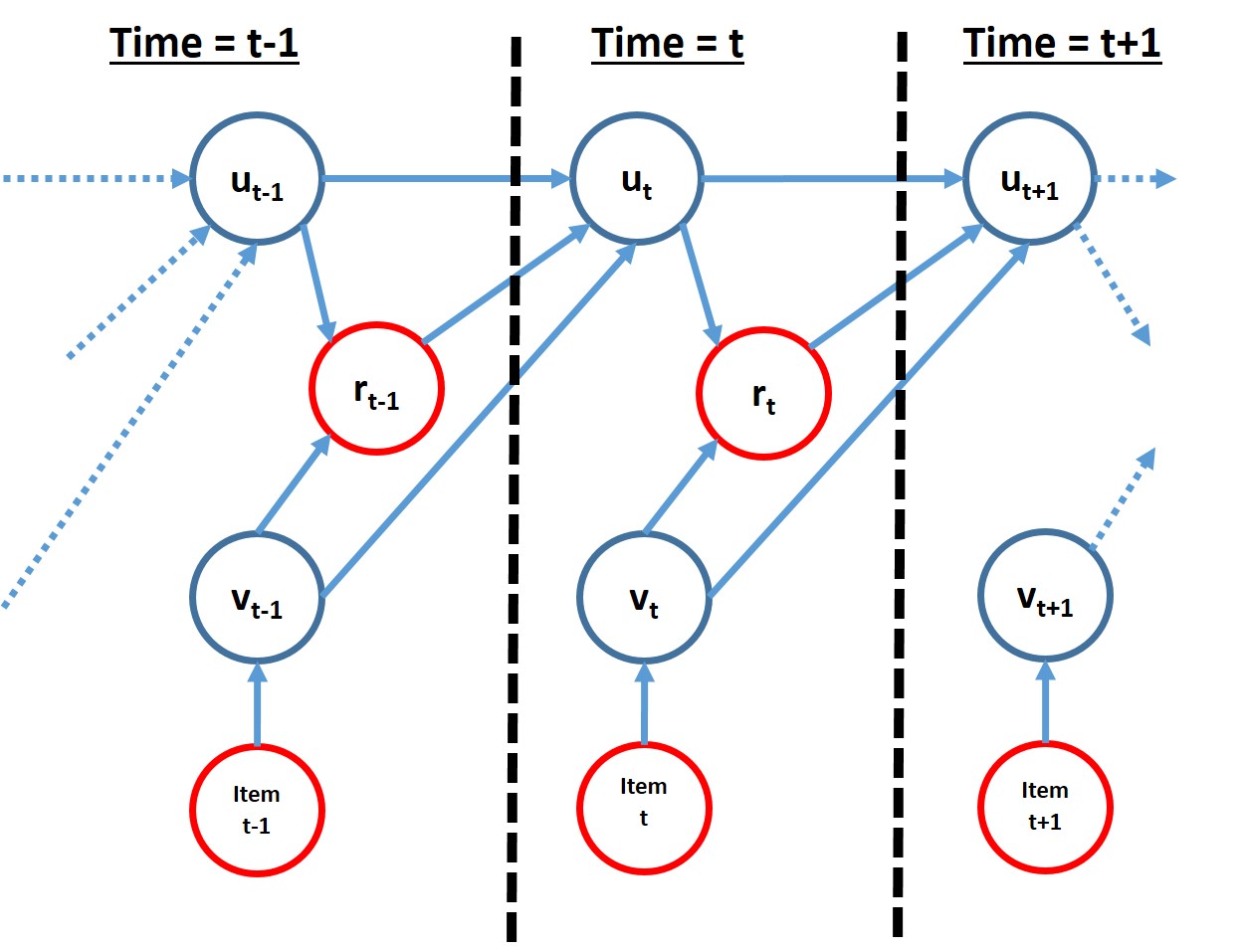}
  \caption{Graphical representation of the suggested perspective. The red circles are known to the learner and the blue ones are the latent representation that requires learning. Arrows symbolize dependency, such that each node (or its mean value) can be calculated using its input nodes.}
  \label{Fig:Scheme}
\end{figure}

The suggested architecture essentially finds important features for estimating the immediate reward and assumes these hold most of the relevant information about the user. However, other architectures can be considered. A general scheme of how we view each trajectory is given in Figure \ref{Fig:Scheme}. Any method or model that learns the vectors $u_t, v_t$ for all user-trajectories is sufficient for applying RL techniques. For example, we can model this scheme as a recurrent neural network \cite{mikolov2010recurrent} where $u_t = g(u_{t-1}, v_{t-1}, r_{t-1})$ for some parameterized function $g$, which is identical for all users and needs to be learned. In this paper we focused on the much more common MF techniques as the main representative of latent space learning. We conclude this section with a few remarks:
\begin{itemize}
	\item It is still possible to include side information and hand-picked features as MF methods normally do, for example by concatenating these to the state vector, or in setting the prior for a Bayesian MF approach.
	\item A semi-MDP is an MDP where each transition lasts a continuous amount of time \cite{sutton1999between}. A semi-MDP can be produced by considering the differences in timestamps instead of their order. For instance, consecutive interactions with the same user that are several minutes apart will be treated differently than interactions parted by weeks. 
\end{itemize}

\section{Off-line Setup} \label{Sec:Off-line}
In this scenario, suppose we are providing recommendation services and we are given data which was acquired through some existing recommender system. The company is looking for ways to increase its revenue over time, and an RL based solution is suggested instead of the solution currently in place. Alas, without reasonable confidence in its improved performance the RL based solution will not be tried out (or A/B tested). This is due to the fact that experiments require many samples because of the low success rate and large catalogs. Following a sub-optimal policy reduces revenue in the short term, and affects the perception of the system by its users, losing credibility and reputation. 

Hence, our goal is to suggest a better policy $\pi_t$ (the target policy) and prove its superiority over the existing behavior policy $\pi_b$, \textbf{without} deploying it on-line. To do so, we pursue the following practice: Perform \textbf{(1) on-policy evaluation} to estimate the value of the behavior policy, \textbf{(2) Estimate the Q-function}, then propose a better policy by \textbf{(3) optimizing over the Q-function}\footnote{A possible alternative is to find the optimal policy directly, also known as direct Policy Search \cite{sutton1999policy}.}, and finally use \textbf{(4) off-policy evaluation} to evaluate the new proposed policy. In the rest of the section we mention specific algorithms we used for this scheme for our empirical evaluation, however, though other options exist for each. We point out the following difficulties:

\begin{itemize}	
	\item \textbf{Curse of dimensionality} -- These four steps force us to use an approximation in a large state space setup. In most real-world venues the number of items (and thus actions) is very large as well so we are compelled to use the state-action featured representation. 
	\item \textbf{Unknown behavior policy} -- Since the data was gathered by another recommender system over a long period, the behavior policy which produced it might be hard to procure, and it might not even be stationary. These problems worsen when the recommendations are detrimental (some items are suggested very rarely or almost exclusively), when the recommender is a mix-product of several recommenders, when non-trivial exploration is employed, if several distinct recommendations are given simultaneously, or when hard rules over-write the initial recommendation\footnote{For example, "Never recommend Whatsapp." since it is so popular recommending it is a waste.}. 
	\item \textbf{Interpretability} --  While we avoid some of the hardships of engineered features, we lose the semantics of the features we procured. Thus, reviewing and analyzing the results is hard. 
\end{itemize}

\subsection{On-policy Evaluation}
First, we need is to estimate the value of the current policy. To do so we use LSTD($0$) \cite{bradtke1996linear}, which essentially finds $\theta_{\pi_b}$ such that $V^{\pi_b}(s) \approx \theta_{ \pi_b}^\top \phi(s)$ through least squares. We use the features found through the process described in Section \ref{Sec:Model}. After finding $\theta_{\pi_b}$, we should ask how to obtain a single value as an estimate. One approach could be to simply take the value of the cold-user, since it represents the LTV expected from a new user. However, when the user interacts with the system many times, the cold start is hardly representative of the average state of users in the system, and indeed the optimization is done over the entire trajectory. Our proposal is to instead sample several states from each trajectory and average over their values. In order to estimate the variance of this process, we use bootstrapping for repeated sampling.

Another possible solution is Monte-Carlo averaging of the cumulative rewards over users: $\hat{J}^{\pi_b} = \frac{1}{m}\sum_{i=1}^m \sum_{t=0}^{T_i} \gamma^t r_t$. This solution provides guarantees in the form of confidence bounds through concentration inequalities. On the given datasets, these bounds were calculated and are orders of magnitude higher than needed for our purposes. Indeed, Monte-Carlo's generalization is weaker and the estimated value is only available for the cold-user.  

\subsection{Q-function Estimation}
Our setup is fitted for LSTDQ($0$) \cite{peters2003reinforcement} which uses linear function approximation for estimating the Q-function of the behavior policy. To use LSTDQ($0$), we need to specify the features of each state-action pair. We suggest to simply concatenate the state and action feature vectors, with an additional constant feature (similarly to \cite{sar2016beyond}). Since the scalar product between these feature vectors is related to the success probability , we suggest concatenating the coordinate-wise product of these vectors as well:
\begin{equation}
\phi(s,a) = (\phi(s), v, \phi(s) \odot v, 1),  \left[ \phi(s) \odot v\right]_i = \left[\phi(s)\right]_i \cdot \left[v\right]_i.
\end{equation}
Ee denote LSTDQ($0$)'s output by $\theta_Q$. As one of our results we show the effect of varying $\gamma$ on $\theta_Q$ for the datasets. If the original recommender system maximized the immediate reward, then we would expect the weights that correspond to $\phi(s) \odot v$ to be higher for lower values of $\gamma$.

\subsection{Q-function Optimization}
Once we have an estimate of the Q-function, ideally we can simply pick the item maximizing it as our policy for one-step policy improvement. However, due to the continuous nature of the state and action spaces, a parametric policy would be more fitting. In addition, if the suggested policy is too distinct from the behavior policy, the  subsequent task of off-policy evaluation might be much harder. Since the behavior policy is unknown, we suggest using the same parametric form for both policies, and optimize over the corresponding parameters. Specifically, we propose $\pi(v|s) \propto \exp(w_\pi^\top \phi(s,v) )$, which also assures every item has positive probability by both policies (also an important property for the off-policy evaluation step). 

For the target policy, we can take $w_{\pi_t}=\alpha \theta_Q$ for some $\alpha >0$ top obtain the softmax policy. As for $w_{\pi_b}$, we need to learn it from the data. Maximizing the likelihood is too computationally expensive due to the denominator: $\sum_a \exp(w_\pi^\top \phi(s,a) )$, so instead\footnote{An additional possible technique is negative sampling \cite{goldberg2014word2vec}.} we suggest dropping the denominator and adding a regularization term weighted by the regularization parameter $\eta$:
\begin{equation}
w_{\pi_b} = \text{argmax}_{w} \left[ e^{-\eta \| w \|^2}  \prod_{h=1}^H \prod_{t=0}^{T_h-1} e^{w^\top \phi(s_{t,h},v_{t,h}) } \right],
\end{equation}
which leads to a simple analytical solution: 
\begin{equation}
w_{\pi_b} = C \sum_{h=1}^H \sum_{t=0}^{T_h-1} \phi(s_{t,h},a_{t,h}),
\end{equation}
for some constant $C>0$. It is now easier to maximize the original likelihood over a scalar parameter $C$, for example through grid search.

Notice that the factors $\pi(v|s) \propto \exp(w_\pi^\top \phi(s,v) )$ still have to be normalized to obtain the policy. Finally, as mentioned before, to reduce the variance of the off-policy estimator the target policy and behavior policy should not differ by much. In order to achieve this we can use any convex combination of $w_{\pi_t},w_{\pi_b}$ to generate an improved policy which is closer to the behavior policy.

\subsection{Off-policy Evaluation}
Following the previous sections, we chose off-policy weighted LSTD(0) \cite{mahmood2014weighted}: 
\begin{equation}
\begin{split}
A &= \sum_{t=0}^T \rho_t \phi  (s_t) \left( \phi(s_t) - \gamma \phi(s_{t+1}) \right)^\top \\
b &= \sum_{t=0}^T \rho_t  \phi (s_t) r_t, \quad\quad\quad \theta_{\pi_t} = A^{-1} b,
\end{split}
\end{equation}
where $\rho_t$ is the importance sampling ratio at time $t$: $\rho_t=\frac{\pi_t(t)}{\pi_b(t)}$. The quantity $\rho_t$ is present in most off-policy evaluation algorithms and usually sets the difficulty of the evaluation problem. When the behavior and target policies are very different, it can get very large, causing the estimate to have high variance, while identical policies lead to the same equations as LSTD($0$). 

In order to get bounds we use bootstrapping. Other methods are included in \cite{theocharous2015ad} - Improved concentration inequalities as suggested in \cite{thomas2015high}, Student's $t$-test and Bias Corrected and accelerated (BCa) bootstrap \cite{efron1987better}, though these are only fitted to the Monte-Carlo estimate.

\section{Experiments} \label{Sec:Exp}
In order to test our approach we applied our architecture on $3$ datasets. The first dataset is the ``Coupon Purchase Prediction" challenge publicly available through Kaggle \cite{CouponDataset}. The second dataset is taken from the proprietary Windows-store-app ``Picks for You" channel. The last data-set is the publicly available Movielens's 1M data set \cite{harper2016movielens}. For all datasets we have ignored available side-information.

Notice that most publicly available recommendation data-sets are not useful for the setup described in this work, as our algorithm requires (User ID, Item ID, Rating, Timestamp) tuples, where the item was proposed by the system. For example, Movielens's dataset is a poor fit since users rate movies at times loosely connected to the times they saw the movie. Still, for reproducibility purposes we preferred to use publicly available datasets, even if we do not expect to see much improvement. 

We compared 2 basic MF algorithms - (1) Solving the basic MF (Equation \ref{Eq:Als}) through Alternating Least Squares (ALS) with random initialization and regularization $\lambda = 0.1$, and (2) SVD applied on the sampled matrix where only the top singular values were taken; Our representation is of size $k=20$ for both methods. To construct the states trajectory per user, for both methods we solved the regularized equation which can be computed using a single ALS iteration:
\begin{equation}
u_t = \left( V \text{diag}(w_t) V^\top + \lambda I_{k\times k} \right)^{-1}   V \text{diag}(w_t) q_t
\end{equation}
where $u_t$ denotes the state at time $t$, $V$ is the item vectors' matrix, $w_t$ is a binary weight vector with $1$'s in elements corresponding to catalog items consumed by the user, and $q_t$ is a vector with the observed ratings. For improved performance, we used the sparse properties of $w_t, q_t$, and Sherman-Morrison's formula \cite{hager1989updating} to iteratively compute the inverse matrix: 
\begin{equation}
\left( V \text{diag}(w_t) V^\top + \lambda I_{k\times k} \right)^{-1} ,
\end{equation}
since every new time step it can increase only by the 1-rank matrix: $v_t v_t^\top$.

We chose $\gamma$ for every set as the maximum likelihood estimate of the drop-out probability: $\gamma = 1- \frac{\text{\# Users}}{\text{\# Samples}}$. In table \ref{Tbl:RMSE} we show the averaged MSE scores over 10-fold cross-validation on each of the data-sets, where we added the constant mean predictor for reference. 

\begin{figure}[!h] 
  \centering
    \includegraphics[width=0.42\textwidth]{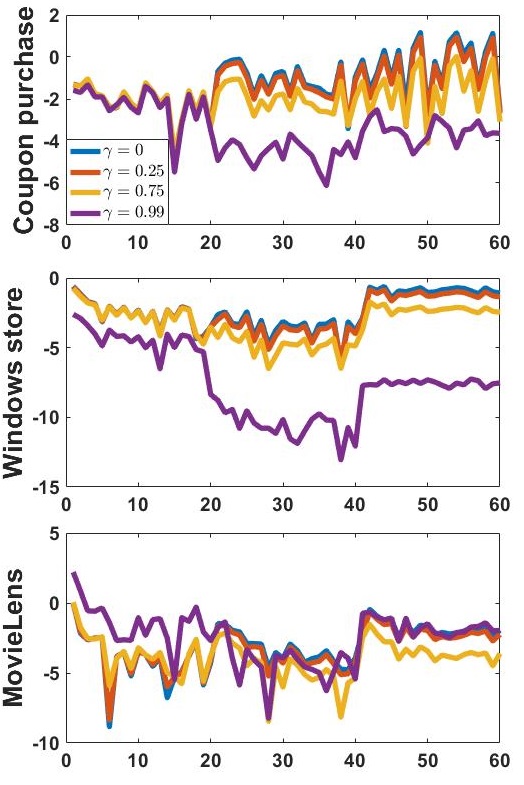}
  \caption{Log-absolute values of the feature's weights estimated by LSTDQ(0). Features $1$-$20$ relate to the state (user), features $21$-$40$ relate to the recommended item and features $41$-$60$ are the coordinate-wise product of the two.}    
  \label{Fig:LSTDQ}
\end{figure}

Next, we provide two interesting plots on the Q-function estimate. In Figure \ref{Fig:LSTDQ} we show the log-absolute value of the weight vector for each of the $60$ features obtained via the SVD approach for varying values of $\gamma$ (the last feature relating to a free parameter was used for normalizing the vector). We expect the last $40$ features relating to the current action to have decreasing effect for larger $\gamma$'s.

\begin{figure*}  
  \centering
    \includegraphics[width=0.9\textwidth]{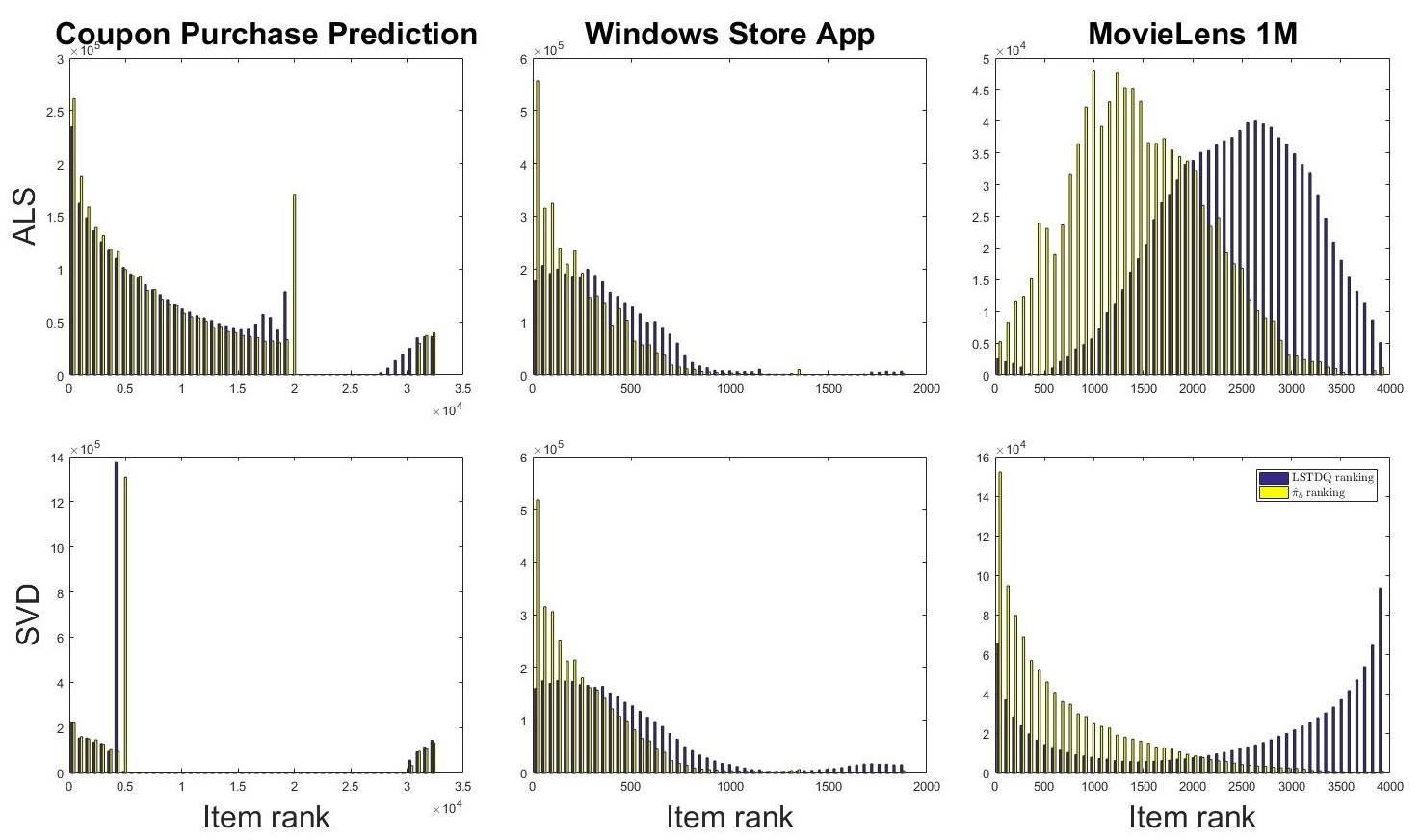}
  \caption{Rankings of the recommended items.}    
  \label{Fig:Ranks}
\end{figure*}

In Figure \ref{Fig:Ranks} we show a histogram of the rankings of the proposed items by both the one-step improvement of the Q-function, and the estimated behavior policy. If we assume the behavior policy aimed to optimize performance, we can expect the Q-function based ranking of the items that were actually offered to be relatively high. Moreover, we expect the recommended item to be highly ranked according to the estimated behavior policy.

Finally, we perform bootstrapping sampling for the on-policy and off-policy evaluation (Table \ref{Tbl:Value}). For the SVD representation which showed better prediction results, we have also used off-policy evaluation to find the estimate of the myopic policy $\pi_m$ (which can be found similarly to $\pi_t$ using LSTDQ with $\gamma=0$). To further evaluate our results, we used one-sided Wilcoxon signed rank test \cite{gibbons2011nonparametric} on the SVD results to test which approach yields higher value, the p-values are reported on the same table for every two policies (all values but one were lower than $10^{-16}$).

\begin{table}[h!]
	\centering
	\begin{tabular}{ | c | c | c | c |} 
		\hline 
		&Coupon & Windows Store & MovieLens \\ 
		\hline
		ALS & $0.076 \pm 10^{-3}$ & $0.34 \pm 10^{-2} $ & $2.99 \pm  0.02$ \\ 
		\hline
		SVD & $0.06 \pm 10^{-4}$ & $0.25 \pm 10^{-3} $ & $1.24 \pm 10^{-3}$\\  
		\hline
		Mean & $0.056\pm 10^{-4}$ & $0.19 \pm 10^{-3} $ & $1.25 \pm 10^{-4}$ \\
		\hline
	\end{tabular}	
\caption{MSE for click/rating prediction.}
\label{Tbl:RMSE}
\end{table}

\begin{table}[!]
	\centering
	\begin{tabular}{ | c | c | c | c |} 
		\hline 
		&Coupon & Windows Store & MovieLens \\ 
		\hline
		$V^{\pi_b}_{\text{ALS}}$ & $9.66 \pm 0.18$ & $51.31 \pm 1.25$ & $16.82 \pm 2.34$ \\ 
		\hline
		$V^{\pi_t}_{\text{ALS}}$ & $11.48 \pm 0.26$ & $50.4 \pm 1.13$ & $17.35 \pm 2.56$\\ 
		\hline		
		$V^{\pi_b}_{\text{SVD}}$ & $9.39 \pm 0.21$ & $51.49 \pm 1.2$ & $21.98 \pm 2.49$\\ 
        \hline		
		$V^{\pi_m}_{\text{SVD}}$ & $9.21 \pm 0.19$ & $52.46 \pm 1.39$ & $22.93 \pm 2.69$\\ 
		\hline		
		$V^{\pi_t}_{\text{SVD}}$ & $ 10.78 \pm 0.64 $ & $54.81 \pm 2.57$ & $26.54 \pm 3.24$\\ 
		\hline		
        \hline
        $p_{\pi_b}^{\pi_m}$ & $1$ & $0$ &$0$ \\        
        \hline
        $p_{\pi_m}^{\pi_t}$ & $0$ & $0$ &$0$ \\        
        \hline
        $p_{\pi_b}^{\pi_t}$ & $0$ & $0$ &$0$ \\        
        \hline
	\end{tabular}	
	\caption{Average value of behavior and target policy for both SVD and ALS based MF. The lower part of the table contains the p--values for one-sided Wilcoxon signed rank test, values lower than $10^{-16}$ were rounded to $0$.}
	\label{Tbl:Value}
\end{table}

\subsection{Coupon Purchase Prediction}
This data set relates to a monetary competition published in the Kaggle website \url{https://www.kaggle.com} in July 2015. The goal of the competition was to improve the recommendations given in the Ponpare coupon site which offers discounts in various markets. The data set is composed of $\sim$3M samples which contain the user ID, coupon ID, whether or not it was purchased and a timestamp. The samples were gathered over a time interval of roughly one year, and include $\sim$23K anonymized users and $\sim$33K items. Before handling the data, we have removed users that had less then 20 interactions with the system or have zero clicks, and we were left with only $\sim$13K users and an average success rate of $0.042$. The discount factor used as is $\gamma = 0.9952$.

As can be seen in Figure \ref{Fig:Ranks}, the behavior and target policy agree more or less on the rank of the items. The fact that there are many badly ranked items implies there might be built-in exploration in the system, or a white list promoting these items. The policy improvement step indeed significantly improves the cumulative discounted reward over the behavior policy, where both SVD and ALS show similar results. Notice that $\pi_t$ also improves over the myopic policy which exhibits worse results than the behavior policy.

\subsection{Windows Store App - "Picks for You"}
This data set relates to the "Picks for You" channel available in the Windows Store App which recommends the user applications to install based on his previous preferences. The recommender system uses a complex and private algorithm for its recommendations, we assumed this system is unknown. The goal in this data set is to increase the Click Through Rate (CTR). For our purposes, we have drawn 2,000 active users over a period of one month, and from these users drew a sample of $\sim$2K active items, which resulted in a total of $\sim$3M samples. No side information available was used here as well. 

In Figure \ref{Fig:Ranks} we can see the rankings of the behavior and target policy. The estimated behavior policy indeed largely agrees with the actual items which were suggested, whereas the target policy rate the recommended items as sub-optimal. Similarly to the Coupon Purchase Prediction challenge, the target policy also recognizes that several very low ranked items were suggested, which might be due to some exploration policy employed by Microsoft, or specific items that have been white-ruled. On this dataset the LTV approach did not show high improvement, but the improvement is statistically significant. 

\subsection{MovieLens}
This data set relates to ratings gathered from the MovieLens website \url{http://movielens.org} and made publicly available by GroupLens Research. They offer various data sizes, where we chose to use the 1M dataset which contains 1 million ratings from 6,000 anonymous users on 4,000 movies. Note that in this dataset the order of the ratings may have been decided by the users (that alternate between movies on a whim) or the system (which recommends the user movies to rate and see). Roughly $0.042$ of the samples were rated, and the ratings were scaled to be between $0$ and $1$, the discount factor is $\gamma = 0.994$. 

In Figure \ref{Fig:Ranks} we can see the rankings of the behavior and target policy. Due to the very high MSE of the ALS MF approach, we suggest ignoring its plot as obviously it failed in grasping the structure of the data. On the SVD plot we can see that indeed the rated items have a higher ranking by the behavior policy, however there is a low correlation between the rank of the item and the fact that it was rated. This corroborates our assumption that items were not rated based on the user preference (so users did not chose to rate only movies they liked), but much more arbitrarily. Figure \ref{Fig:LSTDQ} also supports this claim as varying values of $\gamma$ have low effect on the intensity of the feature weights in the Q-function estimate, suggesting the behavior policy did not try to maximize ratings. 

\section{Summary} \label{Sec:Summary}

Our paper discusses the practical aspects of enhancing a currently operating recommender system with the look-ahead qualities enabled by the field of reinforcement learning. To tackle the difficult problem of feature selection, we proposed to use existing matrix factorization methods. While adding some complexity to the problem, we have shown there can be improvement over time even with the most basic factorization algorithms on relatively small sample datasets. 

Practically, there is much that can be done to improve the empirical results - using other MF or RL algorithms with parameters sweeping, learning on more data, increasing the latent dimension, smart embedding of side information, building a semi-MDP instead of an MDP and so on. However, our goal in this paper was not perfecting results over one specific data set, but offering an entirely new architecture that enables easy modular hands-free LTV optimization. Subsequently, we used two publicly available data-sets and basic easy-to-implement algorithms in every step.


Our framework was designed from an industrial point of view since most recommender systems that process large quantities of users and recommendations belong to data companies. Thus, any change in the production has to be justified, as working systems are rarely tampered with. Subsequently, A/B testing is unavailable not only to outside researchers looking to improve the best known practices, but even to engineers working daily with the product. 

We took this line of thinking into account and the new policy proposed by our algorithm builds upon the existing system, can be chosen arbitrarily close to it, and its improved performance can be proved statistically, or even bounded using the recent literature on off-policy evaluation.

\bibliographystyle{icml2017}
\bibliography{RL4RecSys}

\end{document}